\documentclass[
]{ceurart}

\usepackage{listings}
\begin{document}

\copyrightyear{2025}
\copyrightclause{Copyright for this paper by its authors.
  Use permitted under Creative Commons License Attribution 4.0
  International (CC BY 4.0).}

\conference{CLEF 2025 Working Notes, 9 -- 12 September 2025, Madrid, Spain}

\title{\texorpdfstring{UM\_FHS}{UM	extunderscore FHS} at the CLEF 2025 SimpleText Track: Comparing No-Context and Fine-Tune Approaches for GPT-4.1 Models in Sentence and Document-Level Text Simplification}

\title[mode=sub]{Notebook for the SimpleText Lab at CLEF 2025}


\author[1,2]{Primoz Kocbek}[%
orcid=0000-0002-9064-5085,
email=primoz.kocbek@um.si,
]
\cormark[1]

\address[1]{University of Maribor, Faculty of Health Science, Zitna ulica 15, 2000 Maribor, Slovenia}
\address[2]{University of Ljubljana, Faculty of Medicine, Vrazov trg 2, 1000 Ljubljana, Slovenia}

\author[1,3]{Gregor Stiglic}[%
orcid=0000-0002-0183-8679,
email=gregor.stiglic@um.si,
]
\address[3]{University of Edinburgh, Usher Institute, 5-7 Little France Road, Edinburgh EH16 4UX, UK}


\cortext[1]{Corresponding author.}
\begin{abstract}
This work describes our submission to the CLEF 2025 SimpleText track Task 1, addressing both sentence- and document-level simplification of scientific texts. The methodology centered on using the gpt-4.1, gpt-4.1-mini, and gpt-4.1-nano models from OpenAI. Two distinct approaches were compared: a no-context method relying on prompt engineering and a fine-tuned (FT) method across models. The gpt-4.1-mini model with no-context demonstrated robust performance at both levels of simplification, while the fine-tuned models showed mixed results, highlighting the complexities of simplifying text at different granularities, where gpt-4.1-nano-ft performance stands out at document-level simplification in one case. 
\end{abstract}


\maketitle

\section{Introduction}

This paper presents an overview of the University of Maribor, Faculty of Health Sciences submission (UM FHS) to CLEF 2025 at the SimpleText track \cite{simpletext-lncs} for Task 1 Text Simplification: Simplify scientific text \cite{simpletext-1}, where both subtasks sentence- and document-level simplification were performed. Our previous research in the field of healthcare includes different summarization tasks \cite{kocbek2022generating, stiglic2022relevance} and general application of Large Language Models (LLMs) on multiple downstream tasks \cite{kopitar2024using, kocbek2023evolution}.

This work is the continuation of our previous work from TREC 2024 Plain Language Adaptation of Biomedical Abstracts (PLABA) track\footnote{\href{https://bionlp.nlm.nih.gov/plaba2024/}{https://bionlp.nlm.nih.gov/plaba2024/}}, more specifically Task 2 for complete abstract adaptations that consists of end-to-end biomedical abstracts adaptations for the general public using plain language. We used the guidelines the assume that the average literacy level is lower than grade 8 (<K8), i.e. students 13–14 years old, as is recommended by National Institutes of Health (NIH) for written health materials \cite{hutchinson2016examining}.

Our submissions used the most used LLMs from OpenAI at the time of submission, specifically gpt-4.1 family of models, i.e. gpt-4.1, gpt-4.1-mini and gpt-4.1-nano, where we created two approaches per model through their API. One was a no-context approach with only prompt engineering and the second a fine-tuned (FT) approach. Note that we used the OpenAI API with a signed Data Processing Addendum (DPA)\footnote{\href{https://openai.com/policies/row-privacy-policy/}{https://openai.com/policies/row-privacy-policy/}}, which ensures GDPR compliance.

\section{Approach}

\subsection{Data Description}

The dataset for the SimpleText track \cite{simpletext-lncs} of CLEF 2025 for Task 1 \cite{simpletext-1} focuses on improving access to scientific texts. The dataset contains a collection of scientific documents in various domains, annotated with simplifications to facilitate comprehension. The dataset includes metadata such as document titles, abstracts, and full-text content \cite{hutchinson2016examining}.

\subsection{Models Used}

Since the training dataset is public, we decided to use the GDPR compliant version of OpenAI API, more specifically we used gpt-4.1, gpt-4.1-mini and gpt-4.1-nano, for all model version 2025-04-14.

We used fine-tuning (FT) with an appropriate system prompts for Task 1.1 (Appendix~\ref{appendix:a}) and Task 1.2 (Appendix ~\ref{appendix:b}). We used the provided train and validation data for Task 1.1 and Task 1.2 for FT. We only FT for gpt-4.1-mini and gpt-4.1-nano due to cost constrains as well as performance indications, where gpt-4.1-mini outperformed gpt-4.1. We produced 4 FT models, marked as ft. We used the following hyperparameters: epochs 3, batch size 1, LR multiplier 2, random seed 69517706. 

\subsection{Method Description}

We employed the gpt-4.1 family of models—gpt-4.1, gpt-4.1-mini, and gpt-4.1-nano—for both sentence-level (Task 1.1) and document-level (Task 1.2) adaptations. For each task, we designed custom prompt templates consisting of a system prompt and a user prompt (Appendices ~\ref{appendix:c}–\ref{appendix:f}), supplemented with adapted text simplification guidelines (Appendix~\ref{appendix:g}). In Task 1.1, we enforced strict input-output alignment by requiring that the number of generated simplified sentences exactly match the number of input sentences.

We tested both standard (prompt-only) and fine-tuned (FT) variants of the models. Due to cost constraints, the largest model (gpt-4.1) was only used in its base form. For example, one sentence-level FT on gpt-4.1-mini with the proposed training/validation data costs around USD 24 in training tokens as of the time of writing. 

\subsection{Evaluation Metrics}

We runs were evaluated on standard automatic evaluation measures (SARI, BLEU, FKGL, compression We evaluated the runs using standard automatic metrics, including SARI, BLEU, Flesch-Kincaid Grade Level (FKGL), and compression ratio. To enable a more comprehensive assessment, these quantitative results will be complemented with a detailed human evaluation focusing on qualitative aspects of simplification.

The test data for both Task 1.1 (sentence-level) and Task 1.2 (document-level) was derived from Cochrane abstracts and their corresponding plain language summaries, preprocessed using Cochrane-auto \cite{bakker2024cochrane}. This yielded a benchmark subset comprising 37 paired abstracts (587 source sentences) and 37 corresponding simplified summaries (388 sentences). For Task 1.2, we further evaluated performance on a larger dataset of 217 original abstract-summary pairs, following the approach in \cite{devaraj2021paragraph}.

\section{Results}

The reported results are based on two test sets. The first includes 37 Cochrane abstracts aligned with their plain language summaries via Cochrane-auto, comprising 587 sentence pairs. This dataset was used for evaluating both Task 1.1 (sentence-level) and Task 1.2 (document-level) simplification (Tables \ref{tab:eval11} and \ref{tab:eval12}). The second set consists of 217 unaligned abstract-summary pairs, used exclusively for Task 1.2 evaluation (Table~\ref{tab:eval12_dl}).

\begin{table}
  \caption{Evaluation results for Task 1.1 Sentence-level Scientific Text Simplification (37 Cocrane-auto aligned abstract, best five runs)}
  \label{tab:eval11}
  \begin{tabular}{lrrrr}
    \toprule
    \textbf{Model} &\textbf{SARI} &\textbf{BLEU}& \textbf{FKGL}& \textbf{Compression ratio}\\
    \midrule
    Source	&12.03&	20.53&	13.54&	1.00\\
    Reference&	100.00&	100.00&	11.73&	0.56\\
    \midrule
    gpt-4.1-nano&	29.47&	18.46&	11.10&	0.86\\
    gpt-4.1-nano-ft&	/&	/&	/&	/ \\
    gpt-4.1-mini&	43.34&	13.93&	7.46&	0.78\\
    gpt-4.1-mini-ft&	42.83&	20.85&	12.29&	0.71\\
    gpt-4.1	&38.84&	14.04&	8.51&	0.79\\
    \bottomrule
  \end{tabular}
\end{table}

For Task 1.1 (sentence-level simplification), the best-performing model was gpt-4.1-mini, achieving a SARI score of 43.34. Its readability, as measured by FKGL, was below grade 8, aligning well with NIH guidelines for plain language (targeting a K8 level). In contrast, the reference summaries exhibited a readability closer to grade 12 (K12). Notably, the fine-tuned gpt-4.1-nano (gpt-4.1-nano-ft) failed to generate sentence-level outputs for the test set and was therefore excluded from evaluation.

\begin{table}
  \caption{Evaluation results for Task 1.2 Document-level Scientific Text Simplification (37 Cocrane-auto aligned abstract, best five runs))}
  \label{tab:eval12}
  \begin{tabular}{lrrrr}
    \toprule
    \textbf{Model} &\textbf{SARI} &\textbf{BLEU}& \textbf{FKGL}& \textbf{Compression ratio}\\
    \midrule
    Source&	12.03&	20.53&	13.54&	1.00\\
    Reference&	100.00&	100.00&	11.73&	0.56\\
    \midrule
    gpt-4.1-nano&	37.01&	14.74&	9.05&	0.69\\
    gpt-4.1-nano-ft&	43.61&	16.00&	10.63&	0.50 \\
    gpt-4.1-mini&	43.53&	14.11&	7.48&	0.72\\
    gpt-4.1-mini-ft&	42.82&	22.94&	11.93&	0.60\\
    gpt-4.1&	43.83&	18.12&	8.80&	0.67\\
    \bottomrule
  \end{tabular}
\end{table}

\begin{table}
  \caption{Evaluation results for Task 1.2 Document-level Scientific Text Simplification (217 plain language summaries, best five runs)}
  \label{tab:eval12_dl}
  \begin{tabular}{lrrrr}
    \toprule
    \textbf{Model} &\textbf{SARI} &\textbf{BLEU}& \textbf{FKGL}& \textbf{Compression ratio}\\
    \midrule
    Source&	7.84&	10.55&	13.29&	1.00\\
    Reference&	100.00&	100.00&	11.28&	0.72\\
    \midrule
    gpt-4.1-nano&	28.89&	10.35&	9.90&	0.83\\
    gpt-4.1-nano-ft&	/&	/&	/&	/\\
    gpt-4.1-mini&	42.13&	9.52&	7.56&	0.74\\
    gpt-4. 1-mini-ft&	39.16&	11.95&	12.23&	0.67\\
    gpt-4.1&	37.93&	9.46&	8.82&	0.76\\
    \bottomrule
  \end{tabular}
\end{table}

For Task 1.2 (document-level simplification), using the 37 aligned abstracts, gpt-4.1 achieved the highest SARI score (43.83), closely followed by gpt-4.1-nano-ft (43.61). However, in terms of readability, gpt-4.1 better adhered to NIH guidelines with an FKGL of 8.80, compared to 10.63 for gpt-4.1-nano-ft. When evaluated on the larger dataset of 217 unaligned summaries, performance declined across all models. In this setting, gpt-4.1-mini emerged as the top performer, with a SARI of 42.13 and a favorable FKGL of 7.56, closely matching the target K8 level. Other models underperformed on this dataset, and gpt-4.1-nano-ft produced no usable output.

\section{Discussion and Conclusions}

Model selection remains critical in biomedical text simplification tasks, particularly given the varying levels of complexity even across closely related subtasks. Our results show that models may fail to generalize when prompted with strict or complex rule-based instructions, despite producing output of similar length or structure. For example, in our experiments, the fine-tuned gpt-4.1-nano model frequently failed to generate the desired correct number of sentences when constrained by rule-based prompting. 

From a cost-efficiency perspective, FT smaller models appear attractive. At the time of writing, OpenAI pricing per million training tokens is approximately USD 25 for gpt-4.1, USD 5 for gpt-4.1-mini, and USD 1.5 for gpt-4.1-nano. In our setting, training data amounted to 4.8 million tokens at the sentence level and 2.1 million at the paragraph level, yielding FT costs of USD 24 (gpt-4.1-mini, sentence-level) and USD 7.2 (paragraph-level). However, our results show performance deterioration, except in one case at document level in one case and needs further investigation to assess their overall utility.

Interestingly, comparing the gpt-4.1 family aligns with our insights from the TREC 2024 PLABA track, where the best-performing system for end-to-end biomedical abstract adaptation was based on gpt-4o-mini, outperforming the gpt-4o and needs further investigation.

\begin{acknowledgments}
This work was supported by the Slovenia Research Agency [grant numbers N3-0307, GC-0001]; European Union under Horizon Europe [grant number 101159018].
\end{acknowledgments}

\section*{Declaration on Generative AI}
During the preparation of this work, the author(s) used ChatGPT and Gemini in order to: Grammar and spelling check. After using these tool(s)/service(s), the author(s) reviewed and edited the content as needed and take(s) full responsibility for the publication’s content.

\bibliography{ref_simpletext25_umfhs}

\appendix

\section{System prompt for Fine-tuning Task 1.1}
\label{appendix:a}

\begin{verbatim}
You are SimpleText‑GPT, specialised in adapting biomedical sentences into 
plain language for lay readers.
Follow the  NIH guidelines for written health materials: split long sentences 
if helpful; replace or briefly explain jargon; omit non‑essential statistics; 
allow '' when a sentence is irrelevant; carry over sentences that are already 
plain; preserve every fact; add nothing new. 
INPUT = ['<sentence 1>', '<sentence 2>', …, '<sentence N>'] 
OUTPUT = ['<adaptation 1>', '<adaptation 2>', …, '<adaptation N>']

REQUIREMENTS  
• Return ONE Python list with N elements in the same order. CHECK and than CHECK
again that the number of elements is the SAME as in the INPUT.
\end{verbatim}

\section{System prompt for Fine-tuning Task 1.2}
\label{appendix:b}

\begin{verbatim}
You are SimpleText‑GPT, specialised in adapting biomedical sentences into plain 
language for lay readers.
Follow  the  NIH guidelines for written health materials: split long sentences 
if helpful; replace or briefly explain jargon; omit non‑essential statistics; 
allow '' when a sentence is irrelevant; carry over sentences that are already 
plain; preserve every fact; add nothing new.
\end{verbatim}

\section{System prompt for Task 1.1}
\label{appendix:c}

\begin{verbatim}
You are SimpleText-GPT, an expert biomedical text simplifier. Based on NIH 
guidelines for written health materials.

ESSENTIAL RULES  
• Audience Write for readers at about a US 8th-grade level (K8 or smart 
13-14 year old student).  
• Workflow (1) Carry over each sentence exactly as written, (2) decide 
if it should be adapted or omitted, (3) review the whole list for coherence 
while keeping every '' placeholder.  
• Splitting If a sentence contains more than one idea, split it into shorter
sentences inside the same pair of single quotes; never merge content from 
different source items.  
• Omission If a sentence is irrelevant to lay readers (for example, detailed 
measurement methods), output the empty string '' for that element.  
• Jargon Replace professional terms with common words. If no plain synonym 
exists, keep the term once and add a brief parenthetical gloss.  
• Statistics Remove p-values, confidence intervals, and similar numbers unless 
they are essential for understanding.  
• Voice Use active voice when possible.  
• Pronouns Resolve ambiguous pronouns or other references.  
• Subheadings Remove IMRAD labels, such as ‘Background:’, ‘Introduction:’, 
‘METHODS:’, ‘Results:’, ‘Discussion:’ or integrate them into a full sentence.  
• Output Return one **Python list with N elements**—exactly the same number 
of elements as the input list—and nothing else. Double check this.
\end{verbatim}

\section{User prompt for Task 1.1}
\label{appendix:d}

\begin{verbatim}
TASK – Plain-language sentence adaptation (based on NIH guidelines for written
health materials)

INPUT =['SENTENCE_1', 'SENTENCE_2', …, 'SENTENCE_N']

OUTPUT FORMAT → ['ADAPTATION_1', 'ADAPTATION_2', …, 'ADAPTATION_N']

ESSENTIAL RULES  
• Audience Write for readers at about a US 8th-grade level (K8 or smart 
13-14 year old student).  
• Workflow (1) Carry over each sentence exactly as written, (2) decide if 
it should be adapted or omitted, (3) review the whole list for coherence 
while keeping every '' placeholder.  
• Splitting If a sentence contains more than one idea, split it into shorter 
sentences inside the same pair of single quotes; never merge content from 
different source items.  
• Omission If a sentence is irrelevant to lay readers (for example, detailed 
measurement methods), output the empty string '' for that element.  
• Jargon Replace professional terms with common words. If no plain synonym 
exists, keep the term once and add a brief parenthetical gloss.  
• Statistics Remove p-values, confidence intervals, and similar numbers 
unless they are essential for understanding.  
• Voice Use active voice when possible.  
• Pronouns Resolve ambiguous pronouns or other references.  
• Subheadings Remove IMRAD labels, such as ‘Background:’, ‘Introduction:’, 
‘METHODS:’, ‘Results:’, ‘Discussion:’ or integrate them into a full sentence.  
• Output Return one **Python list with N elements**—exactly the same number 
of elements as the input list—and nothing else. Double check this.

INSTRUCTIONS  
1 Produce one list with N elements in the original order.  
2 For each element follow this three-step process:  
   • First: Carry the sentence over unchanged.  SENTENCE_1 → ADAPTATION_1,
   ..., SENTENCE_N → ADAPTATION_N 
   • Second - decide and modify ADAPTATIONS as needed:
     – If it is already plain → leave it as is.  
     – If it is irrelevant → replace with ''.  
     – Otherwise → simplify it (you may split it).  
   • Third: After processing all items, review the entire list for flow and
   pronoun clarity. Also keep every '' element in place.  
3 Double-check (again) that the output list contains N elements and that no
facts have been added or lost. If the number DO NOT match return to point 1 
and re-do all the steps. Repeat until the number MATCH. 
Return **only** the final list.

QUICK EXAMPLES  
• Simplify 'Myocardial infarction is a leading cause of mortality worldwide.
' → 'A heart attack is a major cause of death worldwide.'  
• Carry over 'Metabolism is essential for life.' → 'Metabolism is essential 
for life.'  
• Omit 'Blood pressure was measured with a sphygmomanometer.' → ''  
• Split 'Cardiovascular disease is the leading cause of mortality, and it is 
influenced by genetics as well as lifestyle.' → 'Heart disease is the leading
cause of death. Genetics and lifestyle also influence it.'
\end{verbatim}

\section{System prompt for Task 1.2}
\label{appendix:e}

\begin{verbatim}
You are SimpleText-GPT, an expert biomedical text simplifier. Based on 
NIH guidelines for written health materials.

ESSENTIAL RULES  
• Audience Write for readers at about a US 8th-grade level (K8 or smart 
13-14 year old student).  
• Splitting If a sentence contains more than one idea, split it into 
shorter sentences inside the same pair of single quotes; never merge 
content from different source items.  
• Omission If a sentence is irrelevant to lay readers (for example, 
detailed measurement methods), output the empty string '' for that 
element.  
• Jargon Replace professional terms with common words. If no plain 
synonym exists, keep the term once and add a brief parenthetical gloss.  
• Statistics Remove p-values, confidence intervals, and similar 
numbers unless they are essential for understanding.  
• Voice Use active voice when possible.  
• Pronouns Resolve ambiguous pronouns or other references.  
• Subheadings Remove IMRAD labels, such as ‘Background:’, ‘Introduction:’, 
‘METHODS:’, ‘Results:’, ‘Discussion:’ or integrate them into a 
full sentence.  
• Output Return only the final simplified sentence as string.
\end{verbatim}

\section{User prompt for Task 1.2}
\label{appendix:f}

\begin{verbatim}
TASK – Plain-language sentence adaptation (based on NIH guidelines for 
written health materials)


ESSENTIAL RULES  
• Audience Write for readers at about a US 8th-grade level (K8 or smart 
13-14 year old student).  
• Splitting If a sentence contains more than one idea, split it into shorter
sentences inside the same pair of single quotes; never merge content from 
different source items.  
• Omission If a sentence is irrelevant to lay readers (for example, detailed 
measurement methods), output the empty string ''.  
• Jargon Replace professional terms with common words. If no plain synonym 
exists, keep the term once and add a brief parenthetical gloss.  
• Statistics Remove p-values, confidence intervals, and similar numbers 
unless they are essential for understanding.  
• Voice Use active voice when possible.  
• Pronouns Resolve ambiguous pronouns or other references.  
• Subheadings Remove IMRAD labels, such as ‘Background:’, ‘Introduction:’, 
‘METHODS:’, ‘Results:’, ‘Discussion:’ or integrate them into a full sentence.  
• Output Return only the final simplified sentence as string.

QUICK EXAMPLES  
• Simplify 'Myocardial infarction is a leading cause of mortality worldwide.' → 
'A heart attack is a major cause of death worldwide.'  
• Carry over 'Metabolism is essential for life.' → 'Metabolism is essential 
for life.'  
• Omit 'Blood pressure was measured with a sphygmomanometer.' → ''  
• Split 'Cardiovascular disease is the leading cause of mortality, and it 
is influenced by genetics as well as lifestyle.' → 'Heart disease is the 
leading cause of death. Genetics and lifestyle also influence it.'

\end{verbatim}

\section{Adapted guidelines}
\label{appendix:g}

\begin{verbatim}
These are guidelines for plain text adaptation from medical texts. The 
guidelines also feature level of importance for specific concepts, if a 
word or multiple words are encased "", that means that this concept has 
the highest priority concept and should always be adhered to in plain 
language adaptations, if a word or multiple words are encased in || 
that means a very high priority concept and should be adhered to in 
plain language adaptations except if it contradicts with a "" concept. 
Similarly word or multiple words encased between [] are high priority 
concepts and should be adhered to except if it contradicts "" or []. 
Examples sentences or example words for plain language adaptations are 
provide in the format // // -> // //, where the first in  // // is the 
original and second sentence  in // // the plain language adaptation.

Education level of audience for adapted (target) text: "K8 (8th grade 
level students, schooling age 13 to 14)"  

|Splitting sentences|: if a sentence is long and contains two or more 
complete thoughts, it should be split into multiple sentences that are 
simpler. All such sentences will be entered in the same cell to the right 
of the source sentence, separating them with periods as per usual. 

|Carrying over sentences or phrases|: a sentence or phrase need not be 
paraphrased if it is already understandable for consumers; it can simply 
be carried over as is. Similarly, some sentences may only need one or two 
terms to be substituted, but no syntactic changes made. 

|Ignoring sentences|: if a source sentence is not relevant to consumer 
understanding of the document, it should be ignored, and the cell to the 
right of it left blank, for example: 
1) Sentences that expound on experimental procedures not relevant to 
conclusions, such as 'Blood pressure of study participants was measured 
in mmHg using a sphygmomanometer.', 
2) Adapt (do not ignore) sentences mentioning or implying that “Future 
studies are needed for this topic...” 

|Resolving anaphora|: if pronouns in the source sentence refer to something
in the previous sentence that is necessary for understanding the current, 
replace them with their referents in the target sentence. For example: 
//Cardiovascular disease is the leading cause of mortality.// -> //Heart 
disease is the leading cause of death.//, //It is influenced by genetics 
as well as lifestyle.// -> //Heart disease is influenced by heredity and 
lifestyle.// 

General guidelines: 
1) [Change passive voice to active voice when possible.] Example //A total 
of 24 papers were reviewed// -> //We reviewed a total of 24 papers//, 
2) [If a source sentence contains a subheading, such as Background:, 
Results:,] a) [And is followed by a complete sentence, omit the subheadings,
such as Background:, Results: in the target text], example //Objective: Our 
aim is to evaluate management of foreign bodies in the upper gastrointestinal 
tract.// -> //Our aim is to rate treatment of foreign objects stuck in the 
upper digestive tract.// b) [And is followed by an incomplete sentence, 
convert the partial or incomplete sentence to a complete target sentence 
by folding in the subheading based on context], examples //Objective: To 
evaluate management of foreign bodies in the upper gastrointestinal tract.
// -> //Our objective is to rate treatment of 
foreign objects stuck in the upper digestive tract.//, //Purpose of this 
review: To evaluate management of foreign bodies in the upper gastrointestinal 
tract.// -> //This review’s purpose is to rate treatment of foreign objects 
stuck in the upper digestive tract.//, 
3) "Omit confidence intervals, p-values, and similar measurements." Example:
//The summary odds ratio (OR) for bacteriologic cure rate significantly 
favored cephalosporins, compared with penicillin (OR,1.83; 95% confidence 
interval [CI], 1.37-2.44); the bacteriologic failure rate was nearly 2 times 
higher for penicillin therapy than it was for cephalosporin therapy 
(P=.00004).// -> //Results favored cephalosporins (antibacterial antibiotics)
over penicillin (another antibiotic).// 
4) [If the current target sentence is partially entailed or implied by the 
previous target sentence, still create a adaptation for the current target 
sentence.] Examples: //The summary odds ratio (OR) for bacteriologic cure 
rate significantly favored cephalosporins, compared with penicillin (OR,1.83; 
95% confidence interval [CI], 1.37-2.44); the bacteriologic failure rate was 
nearly 2 times higher for penicillin therapy than it was for cephalosporin 
therapy (P=.00004).// -> //Results favored  cephalosporins (antibacterial
antibiotics) over penicillin (another antibiotic).//, //The summary OR for 
clinical cure rate was 2.29 (95% CI, 1.61-3.28), significantly favoring 
cephalosporins (P<.00001).// -> //Results favored cephalosporins.// 
5) If the current target sentence can be written EXACTLY as the previous 
target sentence, just type “...” (no quotes) for the current target sentence
Note: this is a rare scenario 
6) [Carry over words that are understandable for consumers OR words that 
consumers are exposed to constantly], such as metabolism. Metabolism does 
not need a substitution, synonym, or adjacent definition in the target 
sentence and can be carried over as is. 
7) [Substitute longer, more arcane words for shorter, more common synonyms.] 
Example: //inhibits// -> //blocks//, //assessed// -> //measured// 
8) "Replace professional jargon with common, consumer-friendly terms." 
a) Examples:  //nighttime orthoses// -> //nighttime braces//, 
//interphalangeal joint// -> //finger knuckle//, b) [If there is ambiguity 
in how a term can be replaced, the full publication or other outside sources 
may be used to deduce the intent of the authors], c) [When substituting a 
term, ensure that it fits in with the sentence holistically, adjusting 
the term or sentence appropriately, e.g. to avoid redundancy. Where 
appropriate, pronouns like it or the general 
you in the adapted term can become more specific from the context.] 
9) "If the jargon or a named entity does not have plain synonyms, leave as is 
in the first mention but explain it with parentheses or nonrestrictive 
clauses."
Subsequent mentions of the same named entity by (1) a PRONOUN or (2) its 
SPECIFIC NAME can be replaced with either (1) a more GENERAL REFERENT or 
(2) its SPECIFIC NAME. Example: //Duloxetine is a combined 
serotonin/norepinephrine reuptake inhibitor currently under clinical 
investigation for the treatment of women with stress urinary 
incontinence.// -> //Duloxetine (a common antidepressant) blocks removal
of serotonin/norepinephrine (chemical messengers) and is studied for 
treating women with bladder control loss from stress.//,  
10) "Treat abbreviations similarly as jargon or named entities. If an 
abbreviation does not have plain synonyms, leave as is in the first mention 
but explain it with parentheses or nonrestrictive clauses." Subsequent 
mentions of the same abbreviation by (1) a PRONOUN or (2) its SPECIFIC 
ABBREVIATION can be replaced with either (1) a more GENERAL REFERENT or
(2) its SPECIFIC ABBREVIATION. Example://This chapter covers antidepressants 
that fall into the class of serotonin (5HT) and norepinephrine (NE) reuptake 
inhibitors.// -> //This work covers antidepressants that block removal of
the chemical messengers serotonin (5-HT) and norepinephrine (NE).//

\end{verbatim}

\end{document}